\documentclass{article} 
\usepackage{nips11submit_e,times}
\usepackage{graphicx} 
\usepackage{amsmath}
\usepackage{amssymb}
\usepackage{natbib}
\usepackage{algorithm}
\usepackage{algorithmic}
\usepackage{wrapfig}
\usepackage{url}
\usepackage{subfigure}

\usepackage[compact]{titlesec}
\titlespacing{\section}{0pt}{*0}{*0}
\titlespacing{\subsection}{0pt}{*0}{*0}
\titlespacing{\subsubsection}{0pt}{*0}{*0}

\nipsfinalcopy

\title{Spike-and-Slab Sparse Coding for Unsupervised Feature Discovery}

\author{
Ian J. Goodfellow, Aaron Courville, Yoshua Bengio \\
D\'epartement d'informatique et de recherche op\'erationnelle \\
Universit\'e de Montr\'eal \\
Montr\'eal, QC H3T 1J4 \\
 \texttt{\{goodfeli@iro.,courvila@iro.,yoshua.bengio@\}umontreal.ca}
}

\begin{document}

\maketitle

\vspace{-.3in}
\begin{abstract}
\vspace{-.15in}
  We consider the problem of using a factor model we call {\em spike-and-slab sparse coding} (S3C) to learn features for a classification task. The S3C model resembles both the spike-and-slab RBM and sparse coding. Since exact inference in this model is intractable, we derive a structured variational inference procedure and employ a variational EM training algorithm. Prior work on approximate inference for this model has not prioritized the ability to exploit parallel architectures and scale to enormous problem sizes. We present an inference procedure appropriate for use with GPUs which allows us to dramatically increase both the training set size and the amount of latent factors.

We demonstrate that this approach improves upon the supervised learning capabilities
  of both sparse coding and the ssRBM on the CIFAR-10 dataset. We evaluate our approach's potential for semi-supervised learning on subsets of CIFAR-10. We
  use our method to win the NIPS 2011 Workshop on Challenges In Learning
  Hierarchical Models' Transfer Learning Challenge.
\end{abstract}


\section{The S3C model}

The S3C model consists of latent binary {\em spike} variables $h \in \{0,1\}^{N}$, latent real-valued
{\em slab} variables $s\in\mathbb{R}^{N}$, and real-valued $D$-dimensional visible vector $v\in\mathbb{R}^{D}$ generated according to this process: $\forall i \in\{ 1,\dots, N\}, d \in \{1,\dots,D\},$
{\small \begin{align}
p(h_i=1) = \sigma( b_i ), \ \ p(s_i \mid h_i) = \mathcal{N}( s_i \mid h_i \mu_i, \alpha_{ii}^{-1} ), \ \ \ p(v_d \mid s, h ) = \mathcal{N}( v_d \mid W_{d:} (h \circ s), \beta_{dd}^{-1} )  \label{eq:s3c_model} 
\end{align}
}\normalsize
where $\sigma$ is the logistic sigmoid function, $b$ is a set of biases on
the spike variables, $\mu$ and $W$ govern the linear dependence of $s$ on
$h$ and $v$ on $s$ respectively, $\alpha$ and $\beta$ are diagonal precision matrices
of their respective conditionals, 
and $h \circ s$ denotes the element-wise product of $h$ and $s$.

To avoid overparameterizing the distribution, we constrain the columns of $W$ to have unit norm, as in sparse coding. We restrict $\alpha$ to be a diagonal matrix and $\beta$ to be a diagonal matrix or a scalar.
We refer to the variables $h_i$ and $s_i$ as jointly defining the $i^{\text{th}}$ hidden unit,
so that there are are total of $N$ rather than $2N$ hidden units. The state of a hidden unit
is best understood as $h_i s_i$, that is, the spike variables gate the slab variables.

In the subsequent sections we motivate our use of S3C as a feature discovery algorithm
by describing how this model occupies a middle ground between
sparse coding and the ssRBM. The S3C model avoids many disadvantages that the ssRBM and
sparse coding have when applied as
feature discovery algorithms. 

\subsection{Comparison to sparse coding}

Sparse coding has been widely used to discover features for classification \citep{RainaR2007}.  Recently \citet{Coates2011b}
showed that this approach achieves excellent performance on the CIFAR10
object recognition dataset. 

Sparse coding \citep{Olshausen-97} describes a class of  generative
models where the observed data $v$ is normally distributed given a set of
continuous latent variables $s$ and a dictionary matrix $W$: $v \sim \mathcal{N}(Ws,\sigma
\mathbb{I})$. Sparse coding places a factorial prior on $s$ such as a
Cauchy or Laplace distribution, chosen to encourage the posterior mode of
$p(s\mid v)$ to be sparse. One can derive the S3C model from sparse coding
by replacing the factorial Cauchy or Laplace prior with a spike-and-slab prior.


One drawback of sparse coding is that the latent variables are not
merely encouraged to be sparse; they are encouraged to remain close to 0,
even when they are active. This kind of regularization is not necessarily undesirable,
but in the case of simple but popular priors such as the Laplace prior
(corresponding to an $L_1$ penalty on the latent variables $s$),
the degree of regularization on active units is confounded with the degree of sparsity.
There is little reason to believe that in realistic settings, these two types of
complexity control should be so tightly bound together. The S3C model avoids
this issue by controlling the sparsity of units via the $b$ parameter that determines
how likely each spike unit is to be active, while separately controlling the
magnitude of active uits via the $\mu$ and $\alpha$ parameters that govern
the distribution over $s$. Sparse coding has no parameter analogous to $\mu$ and cannot
control these aspects of the posterior independently.

Sparse coding is also difficult to integrate into a
deep generative model of data such as natural images. 
While \citet{Yu+Lin+Lafferty-2011} and \citet{Zeiler-ICCV2011} have recently shown some success at learning 
hierarchical sparse coding, our goal is to integrate the feature extraction scheme into a
proven generative model framework such as the deep Boltzmann Machine \citep{Salakhutdinov2009}.
Existing inference schemes known to work well in the DBM-type setting are all either sample-based or are based on variational approximations to the model posteriors, while sparse coding schemes typically employ MAP inference. Our use of variational inference makes the S3C framework well suited to integrate into the known successful strategies for learning and inference in DBM models. It is not obvious how one can employ a variational inference strategy to standard sparse coding with the goal of achieving sparse feature encoding. 

\subsection{Comparison to Restricted Boltzmann Machines}

The S3C model also resembles another class of models commonly used for feature discovery: the RBM.
An
RBM \citep{Smolensky86} is an energy-based model defined through an energy
function that describes the interactions between the obversed data
variables and a set of latent variables.  It is possible to interpret the
S3C as an energy-based model, by rearranging  $p(v,s,h)$
to take the form $\exp\{-E(v,s,h)\}/Z$, with the following energy function:

{\small
\vspace{-.15in}
\begin{equation}
E(v,s,h)=\frac{1}{2} \left(v - \sum_{i} W_{i} s_{i} h_{i} \right)^T \beta \left(v - \sum_{i} W_{i} s_{i} h_{i} \right)
+\frac{1}{2}\sum_{i=1}^{N}\alpha_{i}(s_{i}-\mu_{i}h_{i})^{2}\
-\sum_{i=1}^{N}b_{i}h_{i},
\label{eq:s3c_energy}
\end{equation}
\vspace{-.15in}
} 
 
The ssRBM model family is a good starting point for S3C because it has demonstrated both reasonable performance as a feature discovery
scheme and remarkable performance as a generative model \citep{courville+al-2011}.
Within the ssRBM family, S3C's closest relative is a variant of the \mbox{$\mu$-ssRBM}, defined
by the following energy function:

{\small
\vspace{-0.15in}
\begin{equation}
E(v,s,h)=-\sum_{i=1}^{N}v^{T} \beta W_{i}s_{i}h_{i}\
+\frac{1}{2}v^{T} \beta v
+\frac{1}{2}\sum_{i=1}^{N}\alpha_{i}(s_{i}-\mu_{i}h_{i})^{2}\
-\sum_{i=1}^{N}b_{i}h_{i},
\label{eq:ssrbm_energy}
\end{equation}
} 
\vspace{-0.0in}
where the variables and parameters are defined identically to the S3C. Comparison of equations \ref{eq:s3c_energy} and \ref{eq:ssrbm_energy} reveals that the simple addition
of a latent factor interaction term $\frac{1}{2}(h \circ s)^TW^T \beta W (h \circ s)$ to the ssRBM energy
function turns the ssRBM into the S3C model.
With the inclusion of this term S3C moves from an
undirected ssRBM model to the directed graphical model, described in equation (\ref{eq:s3c_model}). 
This change from undirected modeling to directed modeling has three important effects, that we
describe in the following sections.



{\em The effect on the partition function: } The most immediate consequence of the transition to directed modeling is that the
partition function becomes tractable. This changes the nature of learning algorithms
that can be applied to the model, since most of the difficulty in training an RBM
comes from estimating the gradient of the log partition function. The partition function of S3C is also guaranteed to exist
for all possible settings of the model parameters, which is not true of the ssRBM.



{\em The effect on the posterior:} RBMs have a factorial posterior, but S3C
and sparse coding have a complicated posterior due ot the ``explaining away''
effect. This means that for RBMs, features defined by similar basis functions
will have similar activations, while in directed models, similar features will
compete so that only the most relevant feature will remain active. As shown by \citet{Coates2011b}, the sparse Gaussian RBM is not a very good feature
extractor -- the set of basis functions $W$
learned by the RBM actually work better for supervised learning when these
parameters are plugged into a sparse coding model than when the RBM itself
is used for feature extraction. We think this is due to the factorial posterior.
In the vastly overcomplete setting, being able to selectively activate a
small set of features likely provides S3C a major advantage in discriminative capability.




{\em The effect on the prior: } The addition of the interaction term causes S3C to have a factorial prior.
This probably makes it a poor generative model, but this is not a problem for the purpose of feature discovery.



\section{Other Related work}
The notion of a spike-and-slab prior was established in statistics by
\citet{Mitchell1988}. Outside the context of unsupervised feature discovery for supervised,
semi-supervised and self-taught learning, the basic form of the S3C model  (i.e. a spike-and-slab latent factor model) has appeared a number of times in different domains \citep{Lucke+Sheikh-2011,GarriguesP2008,Mohamed+Heller+Ghahramani-2011,Titsias2011}. To this literature, we contribute an inference scheme that scales to the kinds of object classifications tasks that we consider. We outline this inference scheme next.

\section{Variational EM for S3C}


Having explained why S3C is a powerful model for unsupervised feature
discovery we turn to the problem of how to perform learning and inference in
this model.  Because computing the exact posterior distribution is intractable, we derive an 
efficient and effective inference mechanism and a
variational EM learning algorithm. 

We turn to variational EM
\citep{Saul96} because this algorithm is well-suited for models with latent
variables whose posterior is intractable.  It works by maximizing a
variational lower bound on the log-likelihood called the energy
functional \citep{emview}. More specifically, it is a variant of the EM algorithm
\citep{Dempster77} with the modification that in the E-step, we compute a
variational approximation to the posterior rather than the posterior
itself. While our model admits a closed-form solution to the M-step, we found that online
learning with small gradient steps on the M-step objective worked better in practice.
We therefore focus our presentation on the E-step, given in Algorithm ~\ref{e_step_algorithm}.

\begin{figure*}
    \centering
        \includegraphics[width=2.7in]{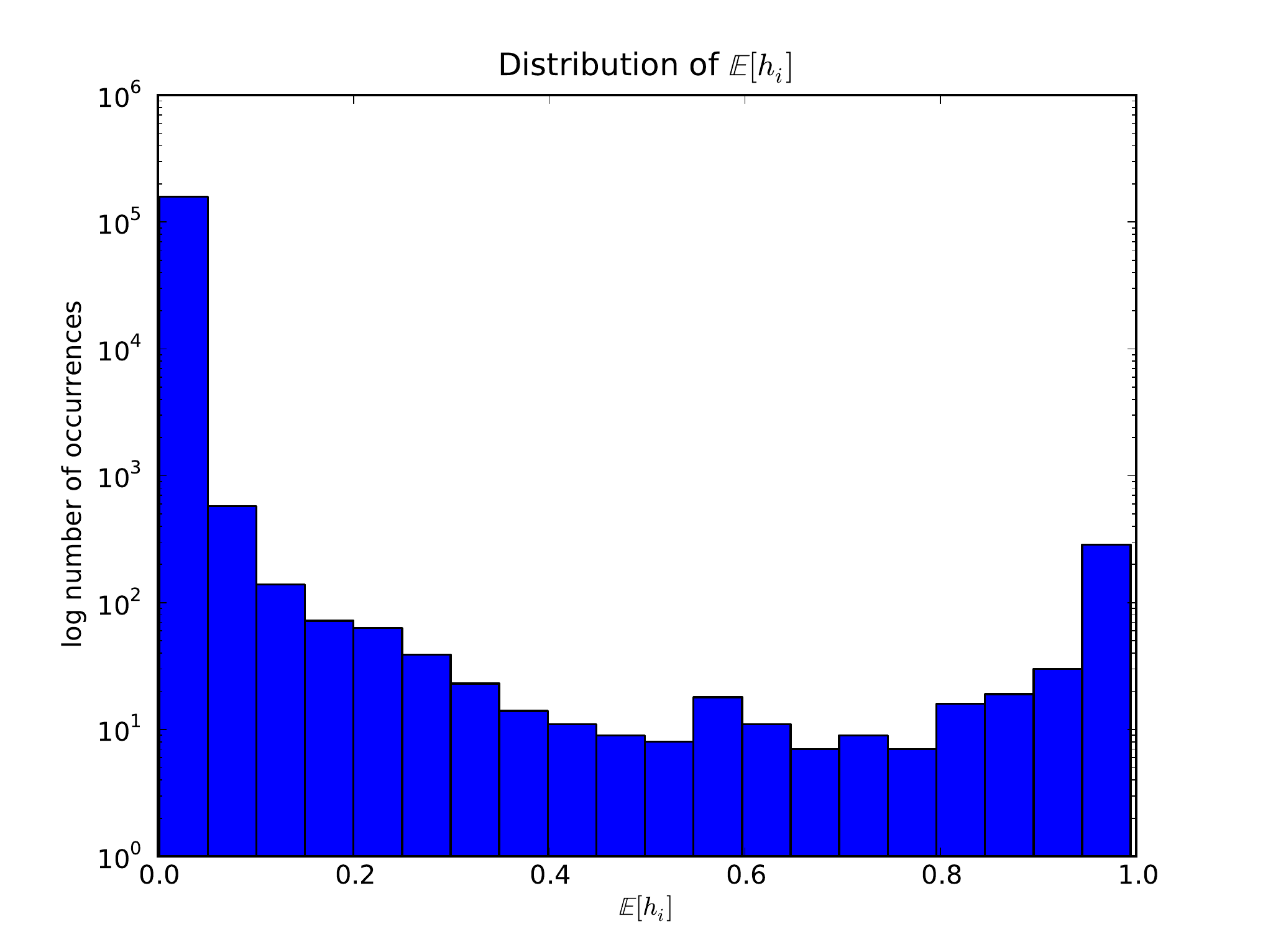}
        \includegraphics[width=2.7in]{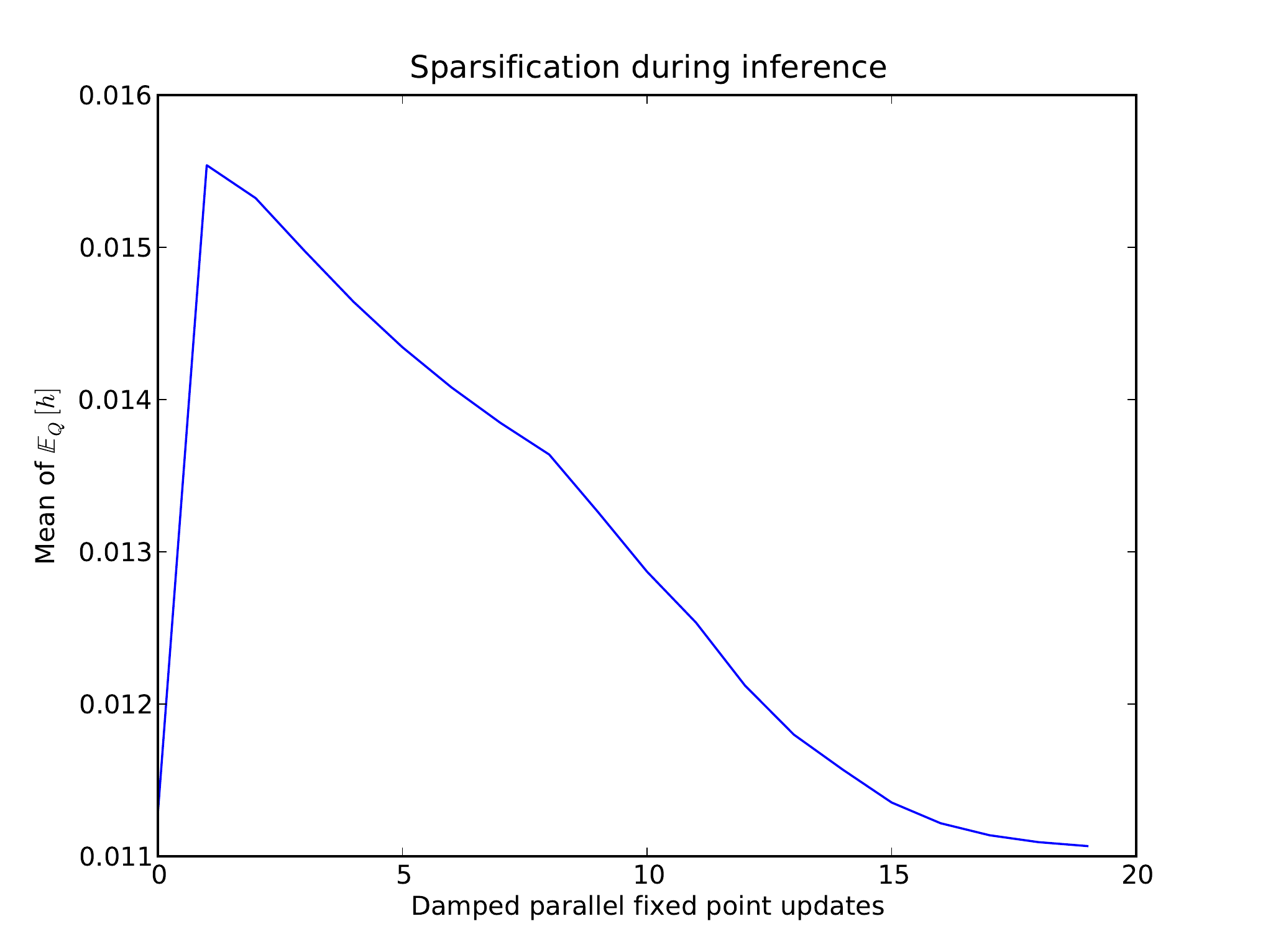}
    \caption{ \small
    (Left) $Q$ imposes a sparse distribution on $h$; $Q(h_i) < .01$ 91.8\% of the time.
   The samples in this histogram are values of $Q(h_i)$ for 1600 different hidden units from a
   trained model applied to 100 different image patches. (Right) The inference procedure sparsifies the representation due to the explaining-away effect. $Q$ is initialized at the prior, which is very sparse.
   The data then drives $Q$ to become less sparse, but subsequent iterations make $Q$ become sparse again.
    }
    \label{hist_h}
\end{figure*}

The goal of the variational E-step is to maximize the energy functional
with respect to a distribution $Q$ over the unobserved variables.
We can do this by selecting the $Q$ that minimizes the Kullback--Leibler divergence:
\vspace{-.05in}
\begin{equation}
\label{min_kl}
 \mathcal{D}_{KL} ( Q(h,s) \Vert P(h,s | v) ) 
\end{equation}
\vspace{-.05in}

where $Q(h,s)$ is drawn from a restricted family of distributions. This
family can be chosen to ensure that $Q$ is tractable.

Our E-step can be seen as analogous to the encoding step of the sparse
coding algorithm.  The key difference is that while sparse coding
approximates the true posterior with a MAP point estimate of the latent
variables, we approximate the true posterior with the distribution $Q$.
We use the family $Q(h,s) = \Pi_i Q(h_i, s_i)$.

Observing that eq.~\eqref{min_kl} is an instance of the Euler-Lagrange equation
\citep{gelfand-book1963}, we find that the solution must take the form 
\begin{align}
\label{family}
Q( h_i ) = \hat{h}_i,\ Q( s_i \mid h_i ) = \mathcal{N} ( s_i \mid h_i \hat{s}_i ,\  (\alpha_i + h_i W_i^T \beta W_i )^{-1} )
\end{align}

\vspace{-.1in}

where $\hat{h}_i$ and $\hat{s}_i$ must be found by an iterative process.
In a typical application of variational inference, the iterative process
consists of sequentially applying fixed point equations that give the
optimal value of the parameters $\hat{h}_i$ and $\hat{s}_i$ for one
factor $Q(h_i,s_i)$ given the value all of the other factors' parameters.
This is for example the approach taken by \citet{Titsias2011} who independently
developed a variational inference procedure for the same problem.
This process is only guaranteed to decrease the KL divergence if applied
to each factor
sequentially, i.e. first updating $\hat{h}_1$ and $\hat{s}_1$ to optimize
$Q(h_1,s_1)$, then updating $\hat{h}_2$ and $\hat{s}_2$ to optimize $Q(h_2,s_2)$,
and so on. In a typical application of variational inference, the optimal
values for each update are simply given by the solutions to the Euler-Lagrange
equations.
For S3C, we make three deviations from this standard approach.

Because we apply S3C to very large-scale problems, we need an algorithm that
can fully exploit the benefits of parallel hardware such as GPUs. Sequential
updates across all $N$ factors require far too much runtime to be competetive
in this regime.

We propose a different method that enables parallel updates to all units.
First, we partially minimize the KL divergence with respect to $\hat{s}$.
The terms of the KL divergence that depend on $\hat{s}$ make up a quadratic
function so this can be minimized via conjugate gradient descent. We implement
conjugate gradient descent efficiently by using the R-operator to perform
Hessian-vector products rather than computing the entire Hessian explicitly
\citep{Schraudolph02}. This step is guaranteed to improve the KL divergence
on each iteration.

We next update $\hat{s}$ in parallel, shrinking the update
by a damping coefficient. This approach is not
guaranteed to decrease the KL divergence on each iteration but it is a widely
applied approach that works well in practice \citep{koller-book2009}.

In practice we find that we can obtain a faster algorithm that reaches equally
good solutions by replacing the conjugate gradient update to $\hat{s}$ with a
more heuristic approach. We use a parallel damped update on $\hat{s}$ much like
what we do for $\hat{h}$. In this case we make an additional heuristic modification
to the update rule which is made necessary by the unbounded nature of $\hat{s}$.
We clip the update to $\hat{s}$ so that if
$\hat{s}_{\text{new}}$ has the opposite sign from $\hat{s}$, its magnitude
is at most $\rho \hat{s}$. In all of our experiments we used $\rho=0.5$ but
any value in $[0,1]$ is sensible. This prevents a case where multiple mutually
inhibitory $s$ units inhibit each other so strongly that rather than being
driven to 0 they change sign and actually increase in magnitude.  This case
is a failure mode of the parallel updates that can result in $\hat{s}$
amplifying without bound if clipping is not used.

\begin{figure*}
    \centering
        \includegraphics[width=2.7in]{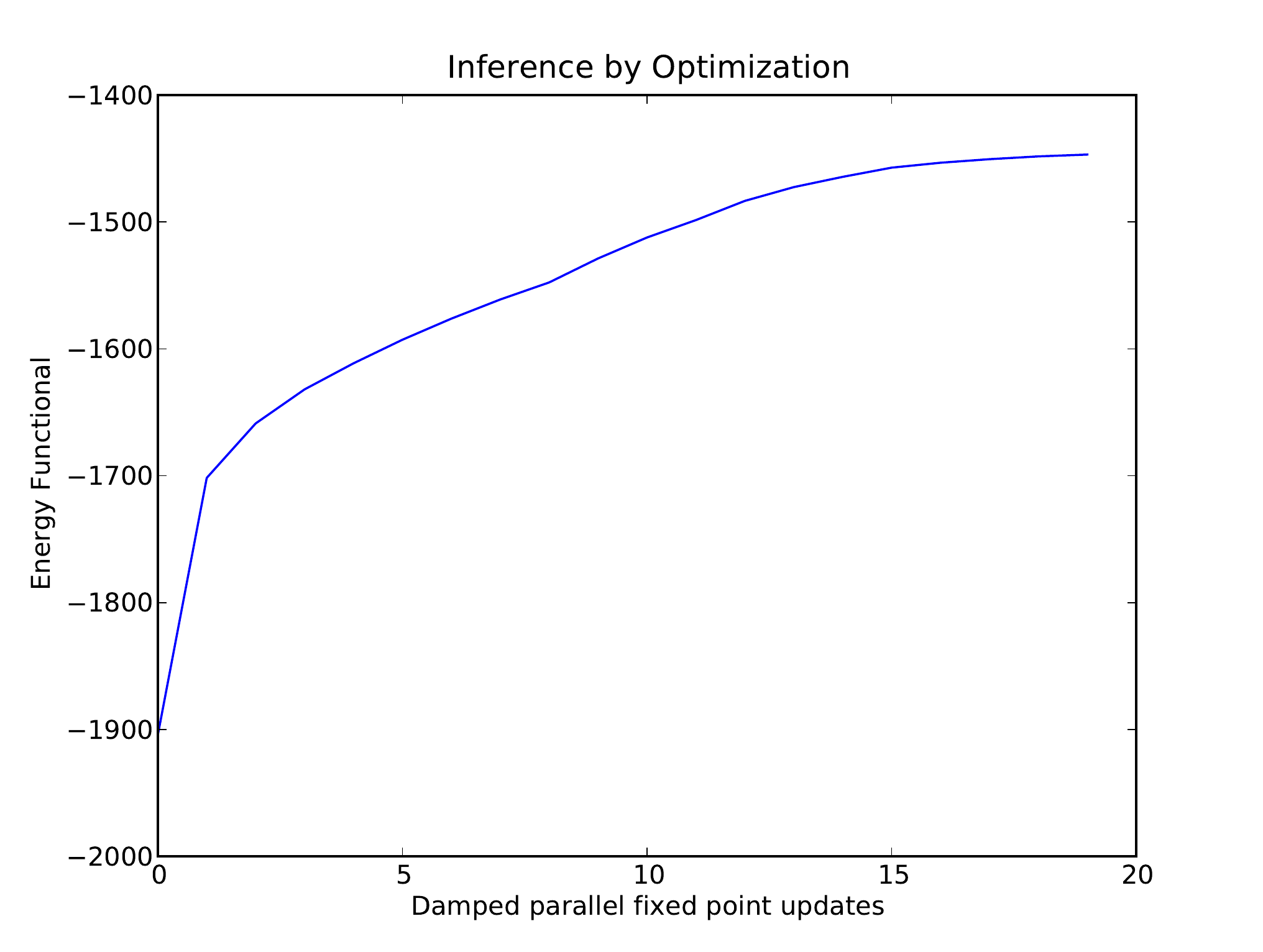}
        \includegraphics[width=2.7in]{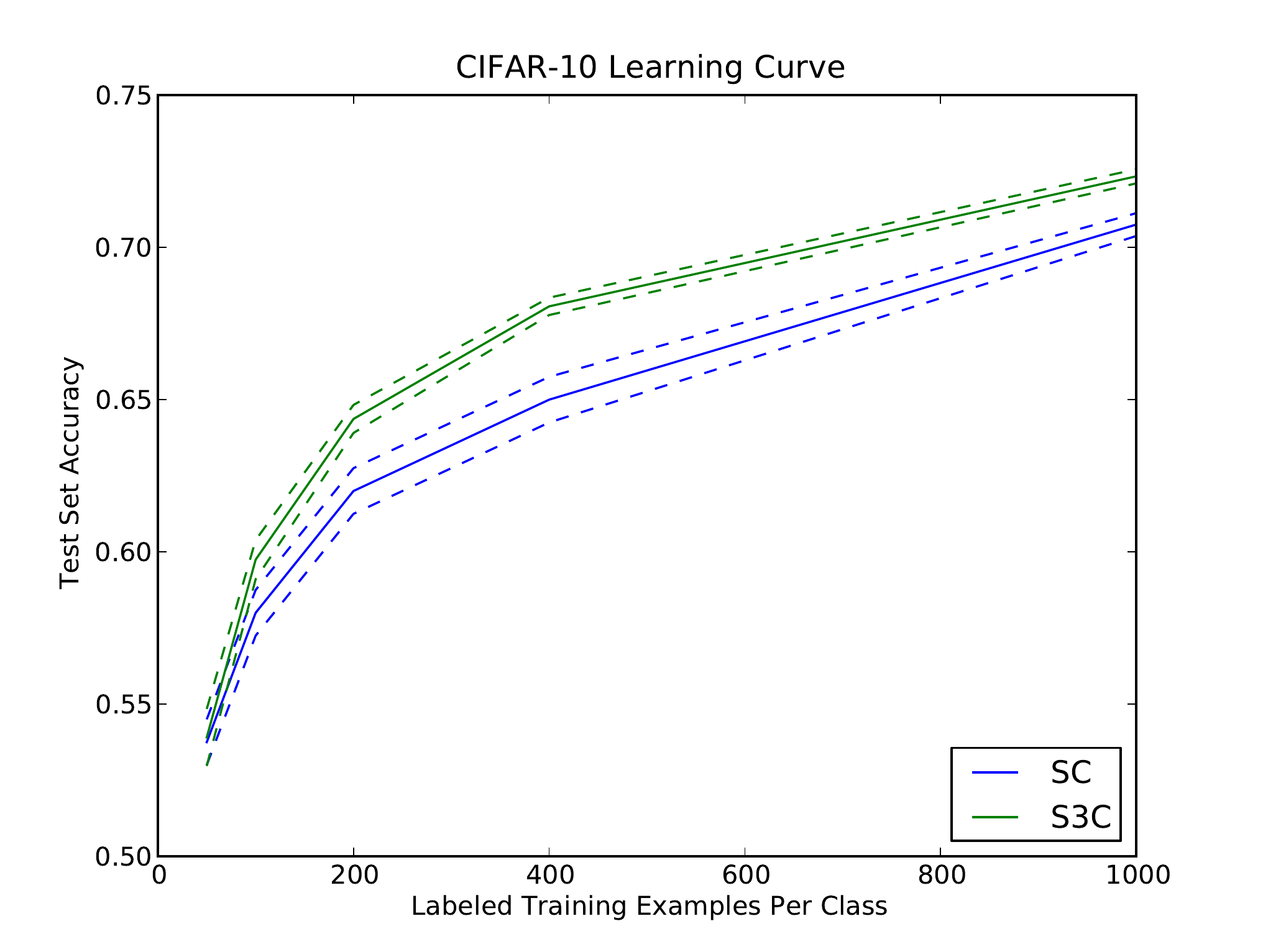}
    \caption{ \small
    (Left) The energy functional of a batch of 5000 image patches increases during the E-step. (Right) Semi-supervised classification accuracy on CIFAR-10.
In both cases the hyperparameters for the unsupervised stage were optimized for performance
on the full CIFAR-10 dataset, not re-optimized for each point on the learning curve.
    }
    \label{energy_functional}
\end{figure*}

We include some visualizations that demonstrate the effect of our E-step.
Figure \ref{hist_h} (right) shows that it produces a sparse representation. Figure
\ref{hist_h} (left) shows that the explaining-away effect incrementally
makes the representation more sparse.
Figure \ref{energy_functional} (left) shows that the E-step increases the energy functional.



\begin{algorithm}[ht!]
\caption[Fixed-Point Variational Inference Algorithm]{{\tt
Fixed-Point Inference}}

\begin{algorithmic}

\tiny{

\STATE Initialize $\hat{h}^{(0)} = \sigma(b)$ and $\hat{s}^{(0)} = \mu$.

\FOR{k=0:K}

\STATE Compute the individually optimal value $\hat{s}_i^*$ for each $i$
simultaneously:

\vspace{-.2in}

\begin{align*}
\hat{s}_i^* &= \frac { \mu_i \alpha_{ii} + v^T \beta W_i - W_i \beta \left[ \sum_{j\neq i} W_j \hat{h}_j \hat{s}_j^{(k)} \right] }
{ \alpha_{ii} + W_i^T \beta W_i }
\label{fixed_point_s}
\end{align*}

\STATE Clip reflections by assigning 

\vspace{-.1in}
\begin{equation*} 
c_i = \rho \text{sign}(\hat{s}^*_i) | \hat{s}_i^{(k)} |
\end{equation*}
\vspace{-.15in}

for all $i$ such that $\text{sign}(\hat{s}^*_i) \neq \text{sign} ( \hat{s}_i^{(k)} ) $
and $| \hat{s}^*_i | > \rho | \hat{s}_i^{(k)} |$, and assigning
$c_i = \hat{s}^*_i$
for all other $i$.

\STATE Damp the updates by assigning

\vspace{-.05in}
\begin{equation*}
\hat{s }^{(k+1)}_i = \eta c + (1-\eta) \hat{s}^{(k)} 
\end{equation*}
where $\eta \in (0,1]$.

\STATE Compute the individually optimal values for $\hat{h}$:

\vspace{-.2in}
\begin{align*}
\hat{h}_i^* &= \sigma \left(
\left( v - \sum_{j\neq i} W_j \hat{s}_j^{(k+1)} \hat{h}_j^{(k)} - \frac{1}{2} W_i \hat{s}_i ^{(k+1)}\right)^T
\beta W_i \hat{s}_i^{(k+1)}
+ b_i \right.\nonumber \\ 
& \left.- \frac{1}{2} \alpha_{ii} ( \hat{s}_i ^{(k+1)}- \mu_i )^2
- \frac {1} {2} \log (\alpha_{ii} + W_i^T \beta W_i )
+ \frac {1} {2} \log ( \alpha_{ii} )
\right) 
\end{align*}

\STATE Damp the update to $\hat{h}$:
\vspace{-.05in}

\begin{equation*}
\hat{h}^{(k+1)} = \eta \hat{h}^* + (1-\eta) \hat{h}^{(k)}
\end{equation*}

\ENDFOR

} 

\end{algorithmic}
\label{e_step_algorithm}
\end{algorithm}





\vspace{-.1in}
\section{Results}

We conducted experiments to evaluate the usefulness of S3C features 
for supervised learning and semi-supervised learning 
on CIFAR-10 \citep{KrizhevskyHinton2009}, a dataset consisting of color
images of animals and vehicles. It contains ten labeled classes, with
5000 train and 1000 test examples per class.

For all experiments, we used the same procedure as \citet{Coates2011b}.
CIFAR-10 consists of $32\times32$ images. We train our
feature extractor on $6\times6$ contrast-normalized and ZCA-whitened patches
from the training set. At test time,
we extract features from all $6\times6$ patches on an image, then average-pool
them. The average-pooling regions are arranged on a non-overlapping grid. Finally,
we train a linear SVM on the pooled features.

\citet{Coates2011b} used 1600 basis vectors in all of their sparse coding experiments.
They post-processed the sparse coding feature vectors by splitting them into the
positive and negative part for a total of 3200 features per average-pooling region.
They average-pool on a $2\times2$ grid for a toal of 12,800 features per image.
We used $\mathbb{E}_Q[h]$ as our feature vector. This does not have a negative part,
so using a $2\times2$ grid we would have only 6,400 features.
In order to compare with similar sizes of feature vectors we used a $3\times3$ pooling
grid for a total of 14,400 features.

\subsection{CIFAR-10}

On CIFAR-10, S3C achieves a test set accuracy
of $78.3 \pm 0.9 $ \% with 95\% confidence 
(or $76.2 \pm 0.9$ \% when using a $2\times2$ grid).
\citet{Coates2011b} do not report test set accuracy for sparse coding with
``natural encoding'' (i.e., extracting features in a model whose
parameters are all the same as in the model used for training) but sparse coding with
different parameters for feature extraction than training achieves an
accuracy of $78.8 \pm 0.9 \%$ \citep{Coates2011b}. Since we have not enhanced our
performance by modifying parameters at feature extraction time these results seem to
indicate that S3C is roughly equivalent to sparse coding for this classification task.
S3C also outperforms ssRBMs, which require 4,096 basis vectors per patch and a $3\times3$
pooling grid to achieve $76.7 \pm 0.9 \%$ accuracy. All of these approaches are close to the state of the art of $82.0 \pm 0.8$ \%,
which used a three layer network \citep{Coates2011c}.

We also used CIFAR-10 to evaluate S3C's semi-supervised learning performance by
training the SVM on small subsets of the CIFAR-10 training set, but using features
that were learned on the entire CIFAR-10 train set. The results, summarized in 
Figure \ref{energy_functional} (right) show that S3C is most advantageous for medium amounts
of labeled data. S3C features thus include an aspect of flexible regularization--
they improve generalization for smaller training sets yet do not cause underfitting on
larger ones.

\section{Transfer Learning Challenge}

For the NIPS 2011 Workshop on Challenges in Learning Hierarchical Models \citep{NipsWorkshop11Hierarchical},
the organizers proposed a transfer learning competition. This competition used
a dataset consisting of 32 $\times$ 32 color images, including 100,000 unlabeled
examples, 50,000 labeled examples of 100 object classes not present in the test
set, and 120 labeled examples of 10 object classes present in the test set.
The test set was not made public until after the competition. We chose to disregard
the 50,000 labels and treat this as a semi-supervised learning
task. We applied the same approach as on CIFAR-10 and won the competition, with a test
set accuracy of 48.6 \%.

\section{Conclusion}

We have motivated the use of the S3C model for unsupervised feature discovery.
We have described a variational approximation scheme that makes it feasible
to perform learning and inference in large-scale S3C models. Finally, we have
demonstrated that S3C is an effective feature discovery algorithm for 
supervised, semi-supervised, and self-taught learning.

\subsubsection*{Acknowledgements}

This work was funded by DARPA and NSERC. The authors would like to
thank Pascal Vincent for helpful discussions. The computation done
for this work was conducted in part on computers of RESMIQ, Clumeq and
SharcNet. We would like to thank the developers of theano
\citep{bergstra+al:2010-scipy} and pylearn2 \citep{pylearn2}. 



\small
\bibliography{strings,strings-shorter,ml,aigaion}
\bibliographystyle{natbib}

\end{document}